%% file: main.tex
\documentclass[11pt, a4paper, copyright, nonumbering]{map}

\usepackage[justification=justified]{caption} %added 

\definecolor{hidden-draw}{RGB}{20,68,106}
\definecolor{hidden-pink}{RGB}{255,245,247}
\input{resources/packages}

\input{resources/math_macro}
\input{resources/macro}

\input{resources/math_commands}

\usepackage{float}  
\usepackage{mathtools}

\hypersetup{
    colorlinks=true,            %链接颜色
    linkcolor=blue,             %内部链接
    filecolor=magenta,          %本地文档
    urlcolor=cyan,              %网址链接
    citecolor=purple,             % 引用颜色
    pdftitle={Efficient Attention Survey},
    pdfpagemode=FullScreen,
    }

\renewcommand{\today}{}
\newcommandx{\info}[2][1=]{\todo[linecolor=red,backgroundcolor=red!25,bordercolor=red,#1]{#2}}

\title{\centering Efficient Attention Mechanisms for Large Language Models: \\A Survey}

\author{
\small
\textbf{Yutao Sun$^{\star}$, 
Zhenyu Li$^{\star}$, 
Yike Zhang$^{\star}$,
Tengyu Pan$^{\star}$, 
Bowen Dong$^{\star}$,}\\
\small
Yuyi Guo,
Jianyong Wang$^{\dagger}$
\\
        \vspace{0.1in}
        \small
Tsinghua University
    \vspace{-0.35in}

}

\vspace{-0.35in}

\input{sections/abstract}

\begin{document}
% \begin{CJK*}{UTF8}{gbsn}
\maketitle

\let\oldthefootnote\thefootnote

\let\thefootnote\relax\footnotetext{$^\star$~Equal Contribution. ~~$^\dagger$~Corresponding Author(s). }
\let\thefootnote\oldthefootnote

\newpage

\tableofcontents

\newpage

\input{sections/introduction}

\input{sections/linear_attention}

\input{sections/sparse_attention}
\input{sections/hybrid}

\input{sections/conclusion}

\newpage

\bibliography{main.bib}

\newpage
% \appendix
% \input{sections/appendix}

% \end{CJK*}
\end{document}

%% file: resources/packages.tex
\usepackage{natbib}

\usepackage{CJKutf8}

\usepackage{xargs}  

\usepackage{todonotes}  

\usepackage{multirow}

\usepackage{cleveref}

\usepackage{amsmath}
\usepackage{dsfont}

\usepackage{subcaption}

\usepackage{svg}
 \usepackage{ragged2e} 
\usepackage{tikz}
\usepackage{threeparttable}   % threeparttable 环境 + tablenotes
\usepackage{multicol}

\usetikzlibrary{arrows.meta, positioning, calc}

\usepackage{graphicx}

\usepackage[edges]{forest}

%% file: resources/macro.tex
\definecolor{bred}{RGB}{250, 82, 82}
\definecolor{borange}{RGB}{253, 126, 20}
\definecolor{byellow}{RGB}{250, 176, 5}
\definecolor{bgreen}{RGB}{116, 184, 22}
\definecolor{bblue}{RGB}{250, 176, 5}
\definecolor{bindigo}{RGB}{76, 110, 245}
\definecolor{bcyan}{RGB}{59, 201, 219}
\definecolor{bteal}{RGB}{99, 230, 190}

%% file: resources/math_commands.tex
\usepackage{amsmath,amsfonts,bm}

\renewcommand\intercal{{\cramped{{}^\mathsf{T}}}}
\def\eqref#1{equation~\ref{#1}}
\def\1{\bm{1}}

\DeclareMathAlphabet{\mathsfit}{\encodingdefault}{\sfdefault}{m}{sl}
\SetMathAlphabet{\mathsfit}{bold}{\encodingdefault}{\sfdefault}{bx}{n}

%% file: sections/abstract.tex
\begin{abstract}
    \vspace{-0.15in}
Transformer-based architectures have become the prevailing backbone of large language models. However, the quadratic time and memory complexity of self-attention remains a fundamental obstacle to efficient long-context modeling. To address this limitation, recent research has introduced two principal categories of efficient attention mechanisms. Linear attention methods achieve linear complexity through kernel approximations, recurrent formulations, or fast-weight dynamics, thereby enabling scalable inference with reduced computational overhead. Sparse attention techniques, in contrast, limit attention computation to selected subsets of tokens based on fixed patterns, block-wise routing, or clustering strategies, enhancing efficiency while preserving contextual coverage. This survey provides a systematic and comprehensive overview of these developments, integrating both algorithmic innovations and hardware-level considerations. In addition, we analyze the incorporation of efficient attention into large-scale pre-trained language models, including both architectures built entirely on efficient attention and hybrid designs that combine local and global components. By aligning theoretical foundations with practical deployment strategies, this work aims to serve as a foundational reference for advancing the design of scalable and efficient language models.

\end{abstract}
\keywords{Language Model, Linear Attention, Sparse Attention, Hybrid Model}

%% file: sections/introduction.tex
\section{Introduction}

Transformer-based architectures~\cite{transformer} have become the de-facto choice as the backbone of modern Large Language Models (LLMs). Despite their success, the standard self-attention mechanism remains a significant computational bottleneck, with quadratic time and memory complexity with respect to input sequence length. This limitation poses substantial challenges for scaling LLMs to handle increasingly long contexts with both strong performance and high efficiency.

To address this, two major directions have emerged to reduce the time and space complexity of $\mathrm{softmax}$ Attention. The first mechanism is Linear Attention~\cite{linear-transformer,retnet,mamba,gla,fastweight,deltanet}, which seeks to reduce attention complexity by reparameterizing or approximating the softmax attention as linear operations. The second candidate is Sparse Attention~\cite{sparsetransformer,jiang2024minference,tang2024questqueryawaresparsityefficient,nsa,moba}, which restricts attention computation to a subset of the full key space based on fixed or dynamic sparsity patterns. While both approaches aim to improve efficiency, they differ significantly in formulation, design choices, and hardware implications.

This survey provides a comprehensive review of recent developments in Efficient Attention mechanisms, with a dual focus on algorithmic principles and system-level implementation. Based on that, we also study the Pre-trained LLM employing these Efficient Attentions.

We categorize linear attention methods into three major paradigms. First, kernelized linear attention approximates the softmax kernel with inner products in a feature space, achieving linear complexity via random feature maps~\cite{performer,rfa} or fixed positive mappings~\cite{linear-transformer}. Second, recurrent linear attention with forgetting mechanisms introduces position-aware recurrence, enabling long-sequence modeling through data-independent~\cite{retnet} or data-dependent decay~\cite{mamba,gla}, which control how past information fades over time. Third, fast-weight and meta-learning-based formulations reinterpret linear attention as a memory update process optimized online, where models such as DeltaNet~\cite{fastweight,deltanet} and TTT~\cite{ttt,ttt2024} incorporate fast-learning dynamics directly into the state evolution. We also examine hardware-friendly representations of linear attention—including parallel, recurrent, and chunkwise forms—highlighting their respective trade-offs in computational complexity, memory footprint, and compatibility with training or inference workflows.

We classify Sparse Attention into fixed-pattern sparsity, block sparsity, and clustering-based sparsity. Fixed-pattern sparsity adopts static token-level masks such as sliding windows, dilated positions, or designated global tokens, offering simplicity and hardware-friendliness~\cite{sparsetransformer, longformer, streamingllm, longnet}. Block sparsity selects or routes attention at block granularity, either via heuristic scoring~\cite{xu2025xattention,tang2024questqueryawaresparsityefficient,moba}, trainable gating~\cite{seerattention,nsa}, enabling structured memory access and efficient GPU utilization. Clustering-based sparsity organizes key-value pairs using content-based or position-aware grouping methods such as k-means or Local Sensitive Hashing (LSH), facilitating semantically aware retrieval with reduced memory overhead~\cite{reformer,chen2024magicpiglshsamplingefficient,liu2024clusterkvmanipulatingllmkv}.
Finally, we also discuss bidirectional sparse designs extend sparsity patterns to encoder-style models.
These approaches differ in sparsity granularity, selection mechanism, and their alignment with hardware primitives like FlashAttention~\cite{flashattn2}, and collectively represent the foundation for efficient long-context modeling in modern Transformers.

There are recent efforts to integrate efficient attention mechanisms into industry-level Pretrained Language Models. These include both pure efficient architectures—such as linear attention and state-space models, and hybrid designs that combine local and global attention patterns. Models like EAGLE~\cite{rwkv56}, Falcon Mamba~\cite{falcon_mamba} and MiniCPM4~\cite{minicpm4} demonstrate the scalability of purely linear or sparse approaches to the multi-billion parameter scale, offering strong performance with constant-time inference. Meanwhile, hybrid models~\cite{gpt3,gemma3,llama4,minimax01,jamba,yoco,rnope} interleave dense, sparse, and local attention to balance computational efficiency with context modeling capacity, reflecting a growing trend toward compositional, hardware-aware attention designs in modern LLMs.

Our goal is to provide a unified framework for understanding the evolution of attention mechanisms under both algorithmic and hardware constraints, and how these designs are integrated into scalable LLM architectures. By connecting theoretical insights with practical implementations, we hope this survey offers a valuable reference for researchers and practitioners working toward efficient and deployable model design.

This survey is motivated by a pivotal transition in the landscape of Large Language Models. While previous development was predominantly driven by a parameter-scaling paradigm aimed at maximizing capability, the current frontier has shifted toward an efficiency-centric orientation. As demands for million-token contexts and low inference cost, the quadratic time-space complexity of canonical self-attention constitutes the principal bottleneck inhibiting scalability. Consequently, the redesign of attention mechanisms has emerged as the critical frontier for architectural innovation. Synthesizing these developments is essential to bridge the disconnect between theoretical algorithmic design and the practical latencies imposed by modern hardware accelerators, thereby laying the groundwork for the next generation of scalable and commercially viable models.

To structure this survey, we organize the discussion as follows:
\begin{itemize}
  \item Section \ref{sec:linear} introduces Linear Attention, covering its evolution across different model generations, the associated design principles, and implications for hardware implementation.
  
  \item Section \ref{sec:sparse} presents Sparse Attention, categorizing sparsity patterns, analyzing deployment scenarios, and offering practical system-level design recommendations.
  
  \item Section \ref{sec:plm} reviews Pre-trained Language Models that incorporate efficient attention mechanisms, including both uniform efficient architectures and hybrid models that integrate local, sparse, and dense attention.
  
  \item Section \ref{sec:takeaways} and \ref{sec:outlook} offers Takeaways and an Outlook on future directions, discussing open challenges and potential advances in algorithmic and hardware-aligned research.
\end{itemize}

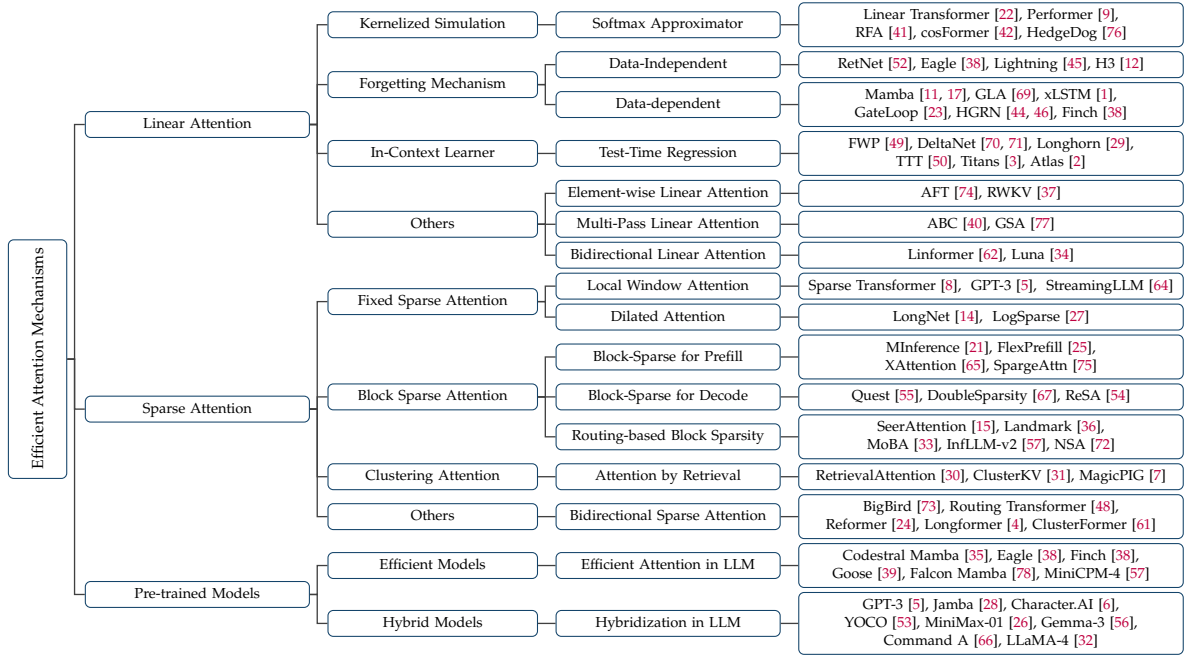
\begin{figure*}[t!]
    \centering
    \resizebox{0.97\textwidth}{!}{
        \begin{forest}
        forked edges,
            for tree={
                grow=east,
                reversed=true,
                anchor=base west,
                parent anchor=east,
                child anchor=west,
                base=center,
                font=\large,
                rectangle,
                draw=hidden-draw,
                rounded corners,
                align=center,
                text centered,
                minimum width=4em,
                edge+={darkgray, line width=1pt},
                s sep=3pt,
                inner xsep=2pt,
                inner ysep=3pt,
                line width=0.8pt,
                ver/.style={rotate=90, child anchor=north, parent anchor=south, anchor=center}
            },
            where level=1{text width=15em,font=\normalsize}{},
            where level=2{text width=14em,font=\normalsize}{},
            where level=3{text width=15em,font=\normalsize}{},
            where level=4{text width=26em,font=\normalsize}{},
            where level=5{text width=20em,font=\normalsize}{},
            [
                \rotatebox{90}{Efficient Attention Mechanisms }
                [
                    Linear Attention, 
                    [
                        Kernelized Simulation, 
                        [
                            Softmax Approximator,
                            [
                                Linear Transformer~\cite{linear-transformer}{, }Performer~\cite{performer}{, }\\RFA~\cite{rfa}{, }cosFormer~\cite{cosformer}{, }HedgeDog~\cite{hedgehog}
                            ]
                        ]
                    ]
                    [
                        Forgetting Mechanism, 
                        [
                            Data-Independent, 
                            [
                                RetNet~\cite{retnet}{, }Eagle~\cite{rwkv56}{, }Lightning~\cite{lightning}{, }H3~\cite{h3}
                            ]
                        ]
                        [
                            Data-dependent,  
                            [
                                Mamba~\cite{mamba,mamba2}{, }GLA~\cite{gla}{, }xLSTM~\cite{xlstm}{, }\\GateLoop~\cite{gateloop}{, }HGRN~\cite{hgrn,hgrn2}{, }Finch~\cite{rwkv56}
                            ]
                        ]
                    ]
                    [
                        In-Context Learner, 
                        [
                            Test-Time Regression
                            [
                                FWP~\cite{fastweight}{, }DeltaNet~\cite{deltanet, gdn}{, }Longhorn~\cite{longhorn}{, }\\
                                TTT~\cite{ttt}{, }Titans~\cite{titans}{, }Atlas~\cite{atlas}
                            ]
                        ]
                    ]
                    [
                        Others,
                        [
                            Element-wise Linear Attention,
                            [
                                AFT~\cite{aft}{, }RWKV~\cite{rwkv}
                            ]
                        ]
                        [
                            Multi-Pass Linear Attention,
                            [
                                ABC~\cite{abc}{, }GSA~\cite{gatedslotattention}
                            ]
                        ]
                        [
                            Bidirectional Linear Attention,  
                            [
                                Linformer~\cite{linformer}{, }Luna~\cite{luna}
                            ]
                        ]
                    ]
                ]
                [
                    Sparse Attention,
                    [
                    Fixed Sparse Attention
                        [
                            Local Window Attention
                            [
                                Sparse Transformer \cite{sparsetransformer}{, }
                                GPT-3 \cite{gpt3}{, }
                                StreamingLLM \citep{streamingllm}
                            ]
                        ]
                        [
                            Dilated Attention
                            [
                                LongNet \cite{longnet}{, }
                                LogSparse \cite{li2019enhancing}
                            ]
                        ]
                    ]
                    [
                        Block Sparse Attention
                        [
                           Block-Sparse for Prefill
                            [
                                MInference \citep{jiang2024minference}{, }FlexPrefill \citep{lai2025flexprefill}{, }\\XAttention \citep{xu2025xattention}{, }SpargeAttn \citep{zhang2025spargeattn}
                            ]
                        ]
                        [
                            Block-Sparse for Decode
                            [
                                Quest \citep{tang2024questqueryawaresparsityefficient}{, }DoubleSparsity \citep{yang2024posttrainingsparseattentiondouble}{, }ReSA \cite{resa}
                            ]
                        ]
                        [
                            Routing-based Block Sparsity
                            [
                                SeerAttention \citep{seerattention}{, }Landmark \citep{mohtashami2023landmark}{, }\\MoBA \citep{moba}{, }InfLLM-v2 \citep{minicpm4}{, }NSA \cite{nsa}
                            ]
                        ]
                    ]
                    [
                        Clustering Attention
                        [
                            Attention by Retrieval
                            [
                                RetrievalAttention \cite{liu2024retrievalattentionacceleratinglongcontextllm}{, }ClusterKV \citep{liu2024clusterkvmanipulatingllmkv}{, }MagicPIG \cite {chen2024magicpiglshsamplingefficient}
                            ]
                        ]
                    ]
                     [
                       Others
                        [
                            Bidirectional Sparse Attention
                            [
                                BigBird~\cite{bigbird}{, }Routing Transformer~\cite{routing}{, }\\Reformer~\cite{reformer}{, }Longformer~\cite{longformer}{, }ClusterFormer~\cite{clusterformer}
                            ]
                        ]
                    ]
                ]
                [
                    Pre-trained Models,  
                    [
                        Efficient Models,
                        [
                            Efficient Attention in LLM
                            [
                                Codestral Mamba~\cite{codestral_mamba}{, }Eagle~\cite{rwkv56}{, }Finch~\cite{rwkv56}{, }\\Goose~\cite{rwkv7}{, }Falcon Mamba~\cite{falcon_mamba}{, }MiniCPM-4~\cite{minicpm4}
                            ]
                        ]
                    ]
                    [
                        Hybrid Models,
                        [
                            Hybridization in LLM,
                            [
                                GPT-3~\cite{gpt3}{, }Jamba~\cite{jamba}{, }Character.AI~\cite{characterai}{, }\\
                                YOCO~\cite{yoco}{, }MiniMax-01~\cite{minimax01}{, }Gemma-3~\cite{gemma3}{, }\\Command A~\cite{rnope}{, }LLaMA-4~\cite{llama4}
                            ]
                        ]
                    ]
                ]
            ]
        \end{forest}}
\captionsetup{justification=centering}
    \caption{Taxonomy of Efficient Attention Mechanisms.}
    \label{fig:taxnomy}
\end{figure*}

%% file: sections/linear_attention.tex
\section{Linear Attention}
\label{sec:linear}

\subsection{Kernelized Linear Attention}
Traditional linear attention methods seek to approximate the softmax-based attention mechanism in a way that scales linearly with sequence length. The core idea is to replace the expensive softmax computation with a \emph{kernel-based} approximation of the attention weights. In standard self-attention, each output is a weighted sum of values $V$ with weights given by a softmax over query-key similarities:
\begin{equation}
\text{Attn}(Q,K,V) = \text{softmax}(QK^\top)V,
\end{equation}
where $Q,K,V \in \mathbb{R}^{L\times d}$ (with $L$ the sequence length and $d$ the model dimension per head). The softmax yields weights $\propto \exp(q_i^\top k_j)$ for query $q_i$ and key $k_j$. Kernelized Linear attention instead finds a feature mapping $\phi(\cdot)$ such that the softmax kernel is approximated by a simple dot-product in an induced feature space: $\exp(q^\top k) \approx \phi(q)^\top \phi(k)$~\cite{kernel-attn}. Given such a $\phi$, one can rewrite attention as:
\begin{equation}
\label{eq:kernel-attn}
O = \frac{\phi(Q)(\phi(K)^\top V)}{\phi(Q)(\phi(K)^\top \mathbf{1})}
\end{equation}
$\phi(\cdot)$ is usually chosen to produce non-negative outputs since $\exp(\cdot)$'s value region is non-negative, a normalization divisor is also applied to mimic the softmax probabilities. This reformulation reduces complexity from $O(L^2d)$ to $O(L d^2)$ (or even $O(Ld)$ with suitable feature dimension reduction), since the expensive $L\times L$ attention matrix is never formed explicitly.

\textbf{Linear Tranformer}~\cite{linear-transformer} replaces the softmax kernel with a fixed positive feature map. In practice they set $\phi(x)=\mathrm{ELU}(x)+1$. $\mathrm{ELU}(\cdot)$ is differentiable in the whole defined region, showing a better performance than naive $\mathrm{ReLU}(\cdot)$ function.

\textbf{Performer}~\cite{performer} introduces FAVOR+ – a Random Features scheme that unbiasedly approximates the softmax kernel.  It samples randomized feature maps $\phi$ so that $E[\phi(Q)\phi(K)^\top] = \exp(QK^\top)$.  This yields a provably unbiased estimator of full softmax attention using only O(N) operations.  In particular, Performers use positive orthogonal random features, which reduce variance in the approximation. 

\textbf{Random Feature Attention}~\cite{rfa} is a linear attention built via Random Fourier Features for the softmax kernel. Similar to Performer, RFA leverages random mapping and triangular activation to approximate $\mathrm{softmax}$. RFA further normalizes the queries and keys before random projection to reduce variance. RFA also has a variant, RFA-Gate, which adds an optional gating mechanism for recency bias.

\textbf{cosFormer}~\cite{cosformer} proposes to use cosine function to approximate $\mathrm{softmax}$.  Since $\cos(a+b)=\cos a \cos b-\sin a \sin b$, cosFormer decomposes the cosine re-weighted attention $S_{ij}=Q'_i K'_j\cos(\frac{\pi}{2}\times\frac{i-j}{M})$ into a linear attention form. $S_{ij}$ represents the attention score between the $i$-th query and $j$-th key. $Q'_i$ and $K'_j$ is a non-negative projection.

\textbf{HedgeDog}~\cite{hedgehog} leverages a spiky kernel $\phi(x)=\exp(Wx+b)$ since they observe that the performance gap between Transformer and Linear Transformer is due to the lack of spiky and monotonic properties. HedgeDog shows a better attention entropy and monotonicity.

\subsection{Linear Attention with Forgetting Mechanism}
A more recent line of work interprets attention through the lens of recurrent neural networks or continuous state-space models. While Traditional Linear Attention is usually position unaware, where the recurrence order does not make influence on the output, modern Linear Attention behaves more like RNNs with state-tracking and hidden memory. Therefore, these models explicitly incorporate recurrence, gating, or state dynamics to handle long sequences with linear complexity. Decay factor is the most important factor to bring forgettimg mechanism.

\subsubsection{Data-Independent Decay}
\label{sec:decay-data-indep}
\textbf{Retentive Networks} (RetNet)~\cite{retnet} introduce a \emph{retention} mechanism that replaces attention with a recurrent-style update using fixed decay coefficients.  In a RetNet layer, each time step $t$ maintains a state vector $s_t$ that aggregates past inputs with exponential forgetting.  The recurrence can be written as
\begin{equation}\label{eq:retnet-recursion}
S_t = \gamma S_{t-1} + k_t^\top v_t
\end{equation}
with $\gamma\in(0,1)$ a learned decay factor (per retention head) and $k_t^\top v_t$ a new contribution from the current token ($v_t$ is a value projection of $x_t$, and $k_t$ is a key projection). The output is then obtained by a linear ``query” projection: $o_t = q_t S_t$. Unrolling \eqref{eq:retnet-recursion} gives an explicit formula for retention:
\begin{equation}\label{eq:retnet-expanded}
o_t =q_tS_t = \sum_{n=1}^{t} \gamma^{t-n}q_tk_t^\top v_t
\end{equation}
which shows that contributions from token $n$ are exponentially decayed by the factor $\gamma^{t-n}$ by the time step $t$. Crucially, $\gamma$ is a \textit{data-independent} decay, which is a fixed parameter for the layer (often one per head in multi-head retention), not a function of the input content. It endows RetNet with an O(1) memory update like an RNN, while still allowing parallel computation during training via an equivalent matrix formulation. (For example, one can show that \eqref{eq:retnet-recursion} is equivalent to a “retention matrix” form $,\mathrm{Retention}(X) = (Q K^\top \odot D),V$, where $D_{t,n}=\gamma^{t-n}$ for $t\ge n$ implements the decay and causal masking.)

RetNet’s retention mechanism shares themes with other data-independent recurrence models. 

\textbf{Eagle}~\cite{rwkv56} improves RWKV design with outer-product memory, which is equivalent to Linear Attention. In RWKV series, the decay factor is parameterized as $\gamma=\exp(-\exp(w))$ where $w$ is a data-independent learnable factor.
% This construction yields a term often denoted $\mathsf{wkv}$ that is mathematically analogous to retention. Aside from this difference, Eagle’s time-mixing module includes an additional learned bias term $u_i$ added to the current $k_t^\top v_t$ (giving that token a special weight) and a sigmoid “receptance” gating that plays a role analogous to a query vector, modulating how much of the decayed state contributes to the output.
In practice, both RetNet and Eagle achieve linear inference scaling with competitive performance, using fixed decays to forget old information. Empirically, RetNet uses a fixed scalar $\gamma$ per head (often each layer has multiple retention heads with different $\gamma$ values, giving a form of multi-scale decay), whereas Eagle uses learnable scalar $w$ to parameterize decay factor.

\textbf{Lightning Attention}~\cite{transnormerllm,lightning} also proposes a linear attention layer augmented with a fixed scalar decay per head to achieve length-invariant computation speed. In Lightning Attention, the hidden state is essentially $s_t = \lambda s_{t-1} + k_t^\top v_t$ for some constant $\lambda$ (with $\lambda$ learned or set by the model), which is in the same spirit as RetNet’s $\gamma$ but optimized for hardware efficiency. 

\textbf{H3}~\cite{h3} introduces a recurrent state-space model~\cite{s4} into linear attention, using learned, data-independent exponential decay via State Space Models (SSM) for key-value outer-product hidden state. While Linear Attention supports efficient chunk-wise training, H3’s reliance on explicit state expansion for SSM computation imposes a practical bottleneck on the head dimension, thereby compromising its expressiveness.

In summary, data-independent decay methods maintain a persistent state that fades over time at a predetermined rate, enabling $O(1)$ recurrence and constant memory per step. They sacrifice some adaptability, which motivates the introduction of data-dependent mechanisms in more recent models.

\subsubsection{Data-Dependent Decay}\label{sec:decay-data-dep}
% TODO: 
While fixed decays offer simplicity and speed, they may under-utilize information from the input stream. Gated or data-dependent approaches make the forgetting factor itself a learned function of the current input. The general form of such a recurrent update is:
\begin{equation}\label{eq:gated-update}
S_t = G_t S_{t-1} + k_t^\top v_t
\end{equation}
where $S_{t-1}$ is the previous state, and $G_t$ is a gating tensor determined by the token $x_t$ . If $G_t$ is close to $0$ in some component, the past state in that component is largely forgotten at time $t$; if $G_t\approx 1$, the past is retained. Unlike the constant $\gamma$ in RetNet, here $G_t$ varies with $t$ via $x_t$. Two notable examples of this strategy in large language model design are Mamba~\cite{mamba,mamba2} and Gated Linear Attention (GLA)~\cite{gla}. 

\textbf{Mamba} is a recurrent state-space model that endows the state decay rates with input dependence. In each Mamba layer, the base state evolution is similar to S4~\cite{s4}, but the state matrix is effectively made dynamic. $G_t$ is a group-wise vector ranging from 0 to 1 as a dynamic forget gate. Which bridges a gap between attention and pure SSMs. Reported empirical results suggest that Mamba2 can achieve competitive or superior performance compared to Transformers of similar or larger scale on specific language modeling tasks, reflecting the potential of data-dependent decay in long-sequence modeling.

\textbf{GLA} directly introduces gating mechanism in the Linear Attention, where a gating function is embedded into a linearized attention layer to improve its expressiveness. GLA modifies retention recurrence by a learnable element-wise forget gate $G_t$.

Beyond these, several other models similarly endow their recurrence with content‑dependent gates.

\textbf{xLSTM}~\cite{xlstm} replaces the standard sigmoid forget gate with an exponential transform of linear gate signals (with normalization), yielding a smooth, input‑conditioned decay on its cell state.

\textbf{GateLoop}~\cite{gateloop} applies head-wise gate on retention, which enables a simple but effective data-dependent decay while maintaining efficient hardware implementation.  

\textbf{HGRN}~\cite{hgrn} introduces gated recurrence in the linear RNNs. HGRN2~\cite{hgrn2} further add state expansion into HGRN framework. State expansion is equivalent to key-value out product in Linear Attention.

\textbf{Finch}~\cite{rwkv56} employs data‑dependent gating on Eagle. Since Eagle is similar to Retention with other orthogonal modifications, Finch also shows deep connection with above models.

In summary, data-dependent decay models augment linear attention or RNN-style architectures with content-based gates that control the flow of information. The results in the paper show that these models can often match or exceed Transformer performance on language tasks, while scaling to very long inputs.

\subsection{Linear Attention as In-Context Learners}
\label{sec:la-icl}
Beyond the efficiency gains offered by linear attention mechanisms, a significant advancement lies in their application to enhance \emph{in-context learning}. This refers to the capability of a model to rapidly adapt or learn from a given prompt without requiring explicit gradient updates to its pre-trained weights. 

While large Transformer models inherently exhibit in-context learning by interpreting the prompt as a form of training data, recent innovations have integrated fast learning rules directly into the attention mechanism, effectively treating sequence processing as an online training process. \textbf{FWP}~\cite{fastweight} establish a formal equivalence between existing linear attention mechanisms and fast weight learning. In the FWP paradigm, a slow neural network learns to program the "fast weights" of another network, often via additive outer products of self-invented key and value patterns.
This section explores several models, including DeltaNet~\cite{deltanet,gdn}, Longhorn~\cite{longhorn}, Test-Time Training (TTT) layers~\cite{ttt,ttt2024}, Titans~\cite{titans}, that exemplify this paradigm of leveraging linear attention as an in-context learner through mechanisms like fast weight updates, viewed through a meta-learning lens.

\begin{table*}[ht]
\centering
\small
\begin{tabular}{@{}ll@{}}
\toprule
\textbf{Method} & \textbf{Update Rule} \\
\midrule
Linear Attention~\cite{linear-transformer} & $S_t = S_{t-1} + k_t v_t^\top$ \\
RetNet~\cite{retnet} & $S_t = \gamma S_{t-1} + k_t v_t^\top$ \\
GLA~\cite{gla} & $S_t = S_{t-1} \mathrm{Diag}(a_t) + k_t v_t^\top$ \\
Mamba~\cite{mamba} & $S_t = \alpha_t S_{t-1} + b_t k_t v_t^\top$ \\
HGRN-2~\cite{hgrn2} & $S_t = S_{t-1} \mathrm{Diag}(a_t) + (1 - a_t) v_t^\top$ \\
DeltaNet~\cite{deltanet} & $S_t = S_{t-1}(1 - \beta_t k_t k_t^\top) + \beta_t k_t v_t^\top$ \\
Gated DeltaNet~\cite{gdn} & $S_t = \alpha_t S_{t-1}(1 - \beta_t k_t k_t^\top) + \beta_t k_t v_t^\top$ \\
TTT~\cite{ttt2024} & $S_t = S_{t-1} - \beta_t \nabla_S \ell(S_{t-1}; k_t, v_t)$ \\
\bottomrule
\end{tabular}
\caption{Update rule among different Linear Attention variants. Each model is a recurrence on matrix memory $S_t$.}
\label{tab:memory_update_rules}
\end{table*}

\subsubsection{Learning Objective}
From a meta-learning perspective, these models define an implicit learning objective that is optimized during inference. Denote $q_t,k_t,v_t$ as the query, key and value at timestep t, the context memory $S_t$ is optimized by the following objective:
\begin{equation}
    \mathcal{L}_t(S)=\frac{1}{2}||f_S(k_t)-v_t||^2
\end{equation}

\textbf{DeltaNet} incorporates the classical delta rule~\cite{fastweight}, where$f_S(k_t)=Sk_t$. Its update rule will be $S_t = S_{t-1} + \eta_t (v_t - S_{t-1}k_t) k_t^\top$, can be derived by minimizing the error between the current memory retrieval $S_{t-1}k_t$ and the new value $v_t$. This signifies a step towards learning the key-value mapping online, effectively refining the memory based on the immediate context.

\textbf{TTT}~\cite{ttt2024} generalizes the meta-learning objective with different modeling architecture:
\begin{equation}
f_S(k_t) =
\begin{cases}
\mathrm{LN}(S\,k_t) + k_t, & \text{TTT-Linear}\\
\mathrm{LN}\bigl(\mathrm{MLP}_S(k_t)\bigr) + k_t, & \text{TTT-MLP}
\end{cases}
\end{equation}

The context network $f_S$ enhances the capability of in-context meta-learning. However, since the gradient of $f_S$ is much more complicated than as a simple linear projection, the online update can not be written as a simple rule.

\paragraph{Batch Update}
Batch update tries to solve the difficulties of the training parallelism when $f_S$ works as a neural network. Usually, context memory is meta‑learned with a batch size of 1, which is not feasible for general TTT models. Instead, analogous to chunk parallelism, TTT treats an entire chunk as a batch. There is no state updates occur within the batch (i.e., $S$ remains consistent). After processing the batch, $S$ is updated once using the aggregated gradients or update signals from all samples in the batch.
This strategy preserves parallel efficiency while accommodating the training requirements of more complex architectures.

\paragraph{Momentum}
\textbf{Titans}~\cite{titans} introduces momentum, which is commonly used in optimization, to strengthen the the capability of the memory update mechanism:

\begin{equation}
    \begin{aligned}
        \mathcal{M}_t&=(1-\alpha_t)\mathcal{M}_{t-1}+S_t \\
        o_t&=q_t\mathcal{M}_t
    \end{aligned}
\end{equation}

The momentum term allows the memory to accumulate information gradually with an exponetial moving average on the state $S$. This can be seen as a form of meta-learning where the update rule itself learns to be more stable and robust over long sequences.

\paragraph{Weight Decay}
Weight decay is another regularization technique in training, corresponding the forgetting mechanism within Linear Attention models. \textbf{Gated DeltaNet}~\cite{gdn} and Titans employs weight decay in its memory update, serving as a learned forget gate to limit the influence of very old or noisy data. It corresponds to the selective state retention mechanisms found in architectures like RetNet~\cite{retnet} and Mamba~\cite{mamba}, where the decay mechanism is proven crucial for language modeling performance:
\begin{equation}
\label{eq:gdn}
    S_{n} = \gamma_nS_{t-1} + \beta_t (v_t - S_{t-1}k_t) k_t^\top 
\end{equation}

In summary, these advancements in linear attention mechanisms are pushing the boundaries of in-context learning by explicitly incorporating meta-learning principles into their architecture. Through fast weight updates, sophisticated memory management techniques, and online learning rules, these models are moving towards a paradigm where the distinction between training and inference becomes increasingly blurred, leading to more efficient and adaptable large language models capable of learning and leveraging knowledge directly from the context.

\subsection{Discussion on Other Designs}
\subsubsection{Element-wise Linear Attention}
\label{sec:element-wise-linear-attention}
\textbf{Attention-Free Transformer}~\cite{aft} leverages a simple weight $\exp(K_{t'}+w_{t,t'})$ instead of $\exp(QK^\top)$:
\begin{equation}
    O_t=\sigma_q(Q_t)\odot\frac{\sum_{t'=1}^t\exp(K_{t'}+w_{t,t'})\odot V_{t'}}{\sum_{t'=1}^t\exp(K_{t'}+w_{t,t'})}
\end{equation}
Where $w_{t,t'}$ is learned as pair-wise position biases. Among the AFT variants, AFT-Simple remove $w_{t,t'}$, achieving linearized inference patterns. Since the product of $K$ and $V$ is element-wise, the recurrent state size is $\mathbb{R}^{d}$ instead of outer-product state $\mathbb{R}^{d\times d}$.

\textbf{RWKV}~\cite{rwkv} leverages decay mechanism on AFT-Simple. Specifically, RWKV improves AFT's position biases with exponential decay $w_{t,i}=-(t-i)w$. The exponential formulation preserves the recurrence property while introducing the position biases.

Element-wise Linear Attention brings strong inference advantage. However, it suffers from the bottleneck of state size, under-performing matrix-based state size. Besides, even though element-wise memory is much fast than outer-product memory, the end-to-end advantage is still marginal since other components occupy more than 95\% latency~\cite{retnet} with outer-product memory.

\subsubsection{Multi-Pass Linear Attention}
\textbf{Attention with Bounded-memory Control} considers Linear Attention as a bounded memory module:
\begin{equation}
    \begin{aligned}
        \tilde{K}_n&=\sum_{i=1}^nK_i \otimes \phi_i,\ \ \tilde{V}_n=\sum_{i=1}^n V_i\otimes\phi_i \\
        O_n&=\mathrm{softmax}(Q_n\tilde{K}_n^\top)\tilde{V}_n
    \end{aligned}
\end{equation}
Where $\tilde{K}_n, \tilde{V}_n$ is the online-updated size-bounded keys and values. In implementation, ABC can be simplified as two-pass Linear Attention.

\textbf{Gated Slot Attention}~\cite{gatedslotattention} further introduces GLA into ABC framework~\cite{abc}. Since $\tilde{K}_n, \tilde{V}_n$ works as an implicit Lienar Attention, GSA improves the update as a gated form:
\begin{equation}
    \tilde{K}_n=\mathrm{Diag}(\alpha_n)\tilde{K}_{n-1}+(1-\alpha_n)\otimes K_n,\ \ \tilde{V}_n=\mathrm{Diag}(\alpha_n)\tilde{V}_{n-1}+(1-\alpha_n)\otimes V_n
\end{equation}
Multi-Pass is an effective way to enhance Linear Attention' expressive ability. However, it also brings additional computation overhead, which makes the architecture design as a trade-off between training efficiency and performance.

\subsubsection{Bidirectional Linear Attention}
Bidirectional attention plays an important role in encoder-style architectures such as BERT~\cite{bert}. The key difference in the linear formulation between unidirectional and bidirectional attention lies in the inference bottleneck and computational pattern. Encoder-only models typically exhibit $O(N^2)$ complexity. Moreover, each token in an encoder-only model has access to global information. As a result, bidirectional linear attention often maintains a constant-length global token pool to reduce complexity while preserving the use of the $\mathrm{softmax}$ function.

For instance, Linformer~\cite{linformer} reduces the number of keys and values to a constant length through an additional matrix projection. 
% Reformer~\cite{reformer} introduces locality-sensitive hashing (LSH) to cluster global tokens into chunks and computes attention within each chunk and the previous one, following the design of Transformer-XL~\cite{transformerxl}. 
Luna~\cite{luna} further extends the Linformer design by encoding the global token pool across model layers.

While bidirectional linear attention is effective for encoder-only architectures, these designs face significant challenges when applied to causal settings as global-pool-based methods tend to be computationally expensive. 
% Cluster-based approaches are nearly infeasible in causal form, and global-pool-based methods tend to be computationally expensive.
Consequently, such architectures are not well-suited for Large Language Models.

\subsection{Hardware Implementation}
\paragraph{The Parallel Representation}

We define the causal linear attention with gated decay as:
\begin{equation}
\begin{aligned}
Q = \phi(X W_Q),\quad K = \phi(X W_K),\quad V = X W_V,\quad \gamma = f_\gamma(X) \\
D_{nm} = 
\begin{cases}
\prod_{i=m+1}^{n} \gamma_i, & n \ge m \\
0, & n < m
\end{cases}, \quad
O(X) = \mathrm{LN}((Q K^\intercal \odot D)V)
\end{aligned}
\end{equation}
where $ W_Q, W_K, W_V \in \mathbb{R}^{d \times d} $, $f_\gamma$ controls the sharpness of decay. The matrix $ D \in \mathbb{R}^{N \times N} $ encodes the causal mask with decay pattern, ensuring uni-directional flow of information.

When the decay is data-independent, $f_\gamma(\cdot)=\mathrm{const}\in(0, 1]$. Note that GroupNorm~\cite{groupnorm} after Linear Attention is already a compulsory component~\cite{retnet}, the explicit devisor of Kernelized Linear Attention in Equation \ref{eq:kernel-attn} is unessential.

The parallel representation is simple and easy to understand but has two shortcomings. First, the parallel form still preserves the $O(L^2)$ complexity, same as $\mathrm{softmax}$ Attention. Second, it's complexity increases when representing the ICL-style Linear Attention in Section \ref{sec:la-icl}.

\begin{figure}[t]
\centering
\begin{subfigure}[b]{0.53\textwidth}
\centering
\includegraphics[width=0.9\textwidth]{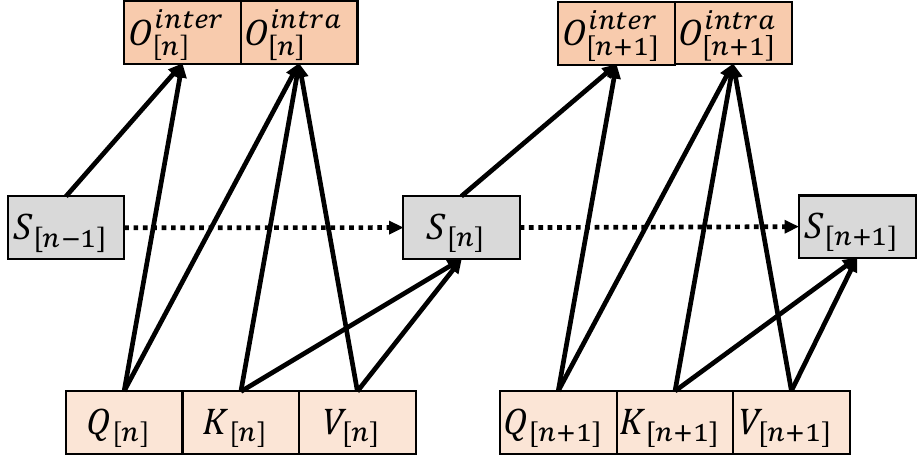}
\caption{Chunkwise representation.}
\label{fig:arch:chunkwise}
\end{subfigure}
\hfill
\begin{subfigure}[b]{0.45\textwidth}
\centering
\includegraphics[width=0.9\textwidth]{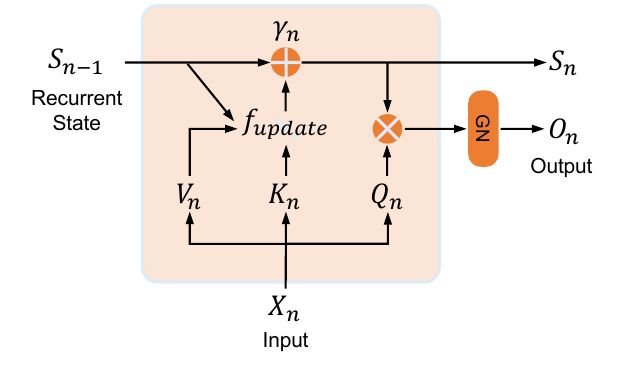}
\caption{Recurrent representation.}
\label{fig:arch:recurrent}
\end{subfigure}

\vspace{1ex}
\caption{Dual form of Linear Attention.}
\label{fig:arch:dual}
\end{figure}

\paragraph{The Recurrent Representation}
The parallel formulation above can be equivalently expressed in a recurrent form for step-wise decoding, as illustrated in \Cref{fig:arch:recurrent}. At each time step $n$, the output is computed as:
\begin{equation}
\begin{aligned}
S_n &= f_{\text{update}}(S_{n-1}, K_n, V_n),\quad O_n = Q_n S_n \\
f_{\text{update}}(S_{n-1}, K_n, V_n) &=
\begin{cases}
\gamma_n S_{n-1} + K_n^\intercal V_n, & \text{Linear Attention with Decay} \\
\gamma_n S_{n-1} + \beta_n (V_n - S_{n-1} K_n) K_n^\top, & \text{ICL-style Linear Attention}
\end{cases}
\end{aligned}
\end{equation}
This recurrent formulation enables efficient auto-regressive generation with constant memory by maintaining a single state vector $S_n$.

While the recurrent representation reduces the computational complexity from $O(L^2)$ to $O(L)$, it incurs substantial memory overhead during training. This is because $S_n$ involves storing outer products of $K_n$ and $V_n$, which is prohibitively expensive for long sequences. As a result, the recurrent form is typically restricted to the decoding stage.

\paragraph{The Chunkwise Recurrent Representation}
The chunkwise representation combines the advantages of linear complexity and hardware-friendly parallelism~\cite{gau,retnet}. As shown in \Cref{fig:arch:chunkwise}, taking decay-style linear attention as an example, given a chunk size $B$, let $x_{[i]}$ denote the $i$-th chunk. Define the cumulative decay within a chunk as:
\begin{equation}
\beta_{(i-1)B+j} = \prod_{k=(i-1)B+1}^{(i-1)B+j} \gamma_k,\quad
D_{[i]}(j,k) = 
\begin{cases}
\frac{\beta_{(i-1)B+k}}{\beta_{(i-1)B+j}}, & j \le k \\
0, & \text{otherwise}
\end{cases}
\end{equation}
The chunk-level memory state $R_i$ is computed as:
\begin{equation}
R_i = K_{[i]}^\intercal (V_{[i]} \odot \frac{\beta_{iB}}{\beta_{[i]}}) + \beta_{iB} R_{i-1}
\end{equation}
and the output for chunk $i$ is given by:
\begin{equation}
O_{[i]} = (Q_{[i]} K_{[i]}^\intercal \odot D_{[i]}) V_{[i]} + (Q_{[i]} R_{i-1}) \odot \beta_{[i]}
\end{equation}

This formulation offers a unified view of recurrence and parallelism: the first term captures intra-chunk dependencies, while the second term propagates inter-chunk memory through a single matrix-vector product. Owing to its efficiency and parallelizability, the chunkwise representation is typically adopted during the training and pre-filling stages.

For ICL-style linear attention, hardware-friendly chunkwise representations have been developed using Householder transformations~\cite{deltanet,gdn}. However, for more sophisticated variants such as TTT and Titans, constructing an explicit chunkwise form remains challenging. Instead, these architectures typically rely on large batch sizes for memory updates, effectively simulating chunkwise computation through a fixed hyperparameter.

Kernel-level optimization is essential for achieving high performance. The widely adopted FLA\cite{fla} provides a Triton-based implementation for many common Linear Attention modules. Alternatively, custom implementations in CUDA or TileLang\cite{tilelang} are also provided by developer, which can be employed for further acceleration.

%% file: sections/sparse_attention.tex
\section{Sparse Attention}
\label{sec:sparse}

Sparse Attention methods employ the inherit sparse property in attention computation and approximate full attention by 

\begin{equation}
\text{Attn}(Q, K, V) = \text{softmax}(QK_{[\mathcal{S}]}^\top) V_{[\mathcal{S}]}
\end{equation}

where \(\mathcal{S}(t)\) is a subset of indices that query vector \(Q(t)\) attend to. Different methods design different selection criteria for \(\mathcal{S}(t)\), taking both selection accuracy and hardware efficiency into consideration. Reduce into sub-linear or linear complexity for prefilling or fixed budget for decoding. 

Conceptually, Sparse Attention differs from Linear Attention in two key dimensions. From a computational perspective, Sparse Attention does not inherently reduce memory footprint, as it still requires storing the full KV cache. Instead, it primarily focuses on accelerating the generation phase by reducing the number of attended tokens. From a performance standpoint, while Linear Attention often involves lossy compression of the global context into a fixed-size state, Sparse Attention maintains the original resolution of selected tokens, allowing its theoretical upper bound to closely approach that of full attention.
% To improve the efficiency of Transformers on long sequences, various \textbf{sparse attention patterns} have been proposed. These approaches reduce the original $O(n^2)$ complexity to near-linear or sub-linear by constraining each query to attend only a subset of key positions, either through predefined or dynamically learned patterns. This section reclassifies existing sparse attention methods according to their attention connectivity patterns. Each pattern is introduced with a brief formula, core idea, and representative methods.

\subsection{Fixed-pattern Sparse Attention}

Fixed-pattern methods represent the earliest and most direct approach within the Sparse Attention paradigm. Because these methods rely on static, pre-defined masks rather than data-dependent routing, they offer the highest degree of hardware friendliness and are the easiest to optimize at the infrastructure level. By leveraging predictable memory access patterns, fixed-pattern designs were the first to be successfully integrated into early long-context Transformers and modern system-level kernels. 
\paragraph{Local Window Attention}
Local-window attention confines each query to interact only with neighbouring tokens inside a fixed sliding window $w$, thus lowering both memory and compute while preserving local context. 

\textbf{Sparse Transformer} \cite{sparsetransformer} first applies local-window (row) attention, where $w$ is close to $\sqrt{N}$, then augments it with an extra column attention that summarizes previous locations  and propagates information globally. \textbf{GPT-3} \cite{gpt3} also adopts a sparse attention pattern similar to that used in Sparse Transformer.

\textbf{StreamingLLM} \cite{streamingllm} found that a large amount of attention score is allocated to initial tokens in the input sequence, which they refer as ``attention sink". They propose a simple Fixed-pattern Attention which keeps only the sink tokens and sliding window tokens. 
For instance, given an input sequence with length \(n\), selected token subset \(\mathcal{S}(t)\) for query token \(q_t\) in StreamingLLM is formulated as
\begin{equation}
\mathcal{S}(t) = \{\, j \mid 0 \le j \le s \  \vee  \ t - w \le j \le t \,\}, \ \forall t\in[1,n]
\end{equation}

where \(s\) is the sink token size and \(w\) is the sliding window size. For better hardware efficiency, StreamingLLM with block granularity \cite{guo2024blocksparse} keeps sink tokens and local tokens in a block-wise manner, enabling efficient memory loading and computation. 

\paragraph{Dilated Attention}

\textbf{LongNet} \cite{longnet} introduces dilated attention as fixed sparse pattern for long context training and inference. Dilated Attention expands the attention field exponentially as the distance grows, thereby reducing the complexity of attention from \(O(L^2)\) to \(O(L)\). Specifically, after dividing input along seqnence dimension into segments with length \(w\), 
dilated sparse index are selected from each segments with an interval \(r\). The selected indices of segment \(i\) is: 

\begin{equation}
\hat{I_i} = [iw, iw+r, iw+2r, ..., (i+1)w-1]
\end{equation}

Sparsified segments \(Q_{\hat{I_i}}, K_{\hat{I_i}}, V_{\hat{I_i}}, \ i \in \{0, 1, ..., \frac{n}{w}\}\) are fed into the attention in parallel, getting attention output \(O\). Combining attention output of different segment sizes and dillation rates \(\{r_i, w_i\}^k\), the final attention is computed as:

\begin{equation}
O = \sum_{i=1}^{K}\alpha_i O|_{r_i, w_i}, \ \alpha_i = \frac{s_i}{\sum_j s_j}
\end{equation}

where \(s_i\) denotes the denominator of the attention softmax for \(O|_{r_i, w_i}\). \textbf{LogSparse} \cite{li2019enhancing} adopts an exponentially sparse attention scheme in which each position attends to only $log N$ tokens, which can be seen an instance of exponentially dilated attention.

% \paragraph{Structured Token-Select Attention}

% \textbf{FastGen} \cite{ge2024modeltellsdiscardadaptive} observed that different attention heads have different structures: there exits head that focus on spectial tokens (such as \texttt{[BOS]} or \texttt{[INST]}), heads that attend to punctuation tokens, heads that has strong locality, and heads that use full attention. 
% Base on this observation, FastGen first conducts lightweight profiling to decide head pattern offline, then adaptively employs sparse attention computation for each head. 
% \begin{equation}
% \mathcal{S}(t) = \{\, j \mid j \in f(K, V, C)\,\}, \ \forall t\in \text{decode tokens}
% \end{equation}
% where \(C\) is a head-level token select policy, \(f\) is a compress function that returns the indices policy \(C\) keeps in the compressed KV Cache. 

\subsection{Block Sparse Attention}

The shift from fixed-pattern to blockwise sparse attention addresses the inherent rigidity of static masks, which fail to capture the dynamic, data-dependent nature of attention in complex tasks. While token-level dynamic selection offers flexibility, it often leads to irregular memory access that severely degrades GPU throughput. Consequently, block-level sparsity emerges as a superior trade-off, which provides the necessary adaptability to focus on task-relevant context while maintaining block-aligned memory access. This structured granularity allows models to leverage high-performance primitives, effectively balancing algorithmic expressiveness with system-level efficiency.

\begin{figure}
    \centering
    \includegraphics[width=0.9\linewidth]{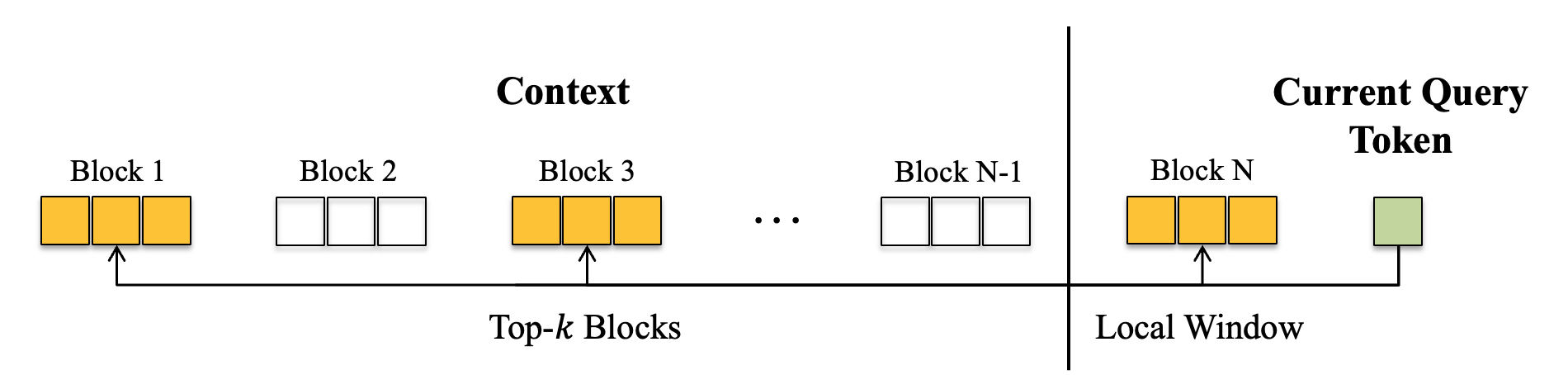}
    \caption{Block-sparse attention: the long sequence is divided into several blocks, and each token attends only to its local window and the top-k related blocks.}
    \label{fig:Block_sparse}
\end{figure}

Given an input sequence with length \(n\) and a block size \(b\), we could divide \(Q, K, V \in \mathcal{R}^{n \times d} \) each into \(\frac{n}{b}\) blocks, each block sized \(b \times d\). 
The goal is to approximate a block-level Mask \(M \in \{0,1\}^{n/b \times n/b}\) used for selecting critical blocks for computation, as illustrated in \Cref{fig:Block_sparse}. 
\begin{equation}
\text{Attn}(Q, K, V)_{i} = \sum_{j=1}^{n/b} M_{ij} \cdot \text{softmax} (Q_i K_j^T ) V_j
\end{equation}
% zyk: 这里 softmax 只包含 mask 保留的块, 另一种形式是

% \[
% \text{Attn}(Q, K, V)_{i} = \text{softmax} (\text{sparse}(Q K^T, i)) V
% \]

Blockwise selection is crucial to achieve efficient computation on modern GPUs.

\subsubsection{Block-Sparse Attention for Prefill}

Methods that use Block-Sparse Attention for prefilling approximate Top-K blocks that cover the majority of attention score with high recall ratio, thus reducing the computation complexity of attention from \(O(L^2)\) to \(O(LK)\).

\begin{equation}
\begin{aligned}
S = \text{softmax}&(Q K^T - c(1-M)) \\
\text{min} \ \ |S&(M) - S_{dense} |
\end{aligned}
\end{equation}

where \(M\) is our block-wise sparse mask defined as above, and \(c\) is a large constant, such as 1e5, ensuring that less important attention weights approaches zero after the softmax computation. 
The objective of the block-sparse attention is to achieve greater speedup with minimal overhead
while retaining as much of the attention weights as possible. 
% Formally, this can be expressed as:

% \begin{equation}
% \begin{aligned}

% \text{min} \ t_{sparse}(M) + t_{overhead}(M)
% \end{aligned}
% \end{equation}

% Given the standard attention weights function: 

\textbf{MInference} \cite{jiang2024minference} observed that there are there patterns in attention weights: Streaming (A Shape) Pattern, Vertical-Slash Pattern and Block-Sparse Pattern. It determines the optimal pattern for each attention head offline and dynamically builds sparse indices based on the assigned pattern during inference.

\textbf{FlexPrefill} \cite{lai2025flexprefill} proposes a context-aware sparse attention mechanism that dynamically adjusts attention patterns and computational budgets in real-time. 

\textbf{XAttention} \cite{xu2025xattention} presents a block-sparse attention framework that utilizes antidiagonal scoring to predict the importance of attention blocks, which could efficiently identify and prune non-essential blocks, achieving high sparsity and substantial computational gains. 

\textbf{SpargeAttn} \cite{zhang2025spargeattn} also employs block-level sparse attention for prefill, which is done through a double-stage online filtering process: the first stage rapidly predicts the attention map to skip certain matrix multiplications, and the second stage applies a softmax-aware filter to further eliminate unnecessary computations.

\subsubsection{Block-Sparse Attention for Decode}

Methods that employ Block-Sparse Attention for decoding dynamically select a subset \(S\) of \(K, V\) vector that contain the most critical tokens for each decoding step, thus reducing memory loading and improving efficiency.

% \[
% \text{Attn}(q, K, V) =  \text{softmax} (q K_{[S]}^T) V_{[S]}
% \]

\textbf{Quest}
\cite{tang2024questqueryawaresparsityefficient} approximate the criticality of each block by calculating an upper bound of attention weights.  For block \(K_i\) we maintain element-wise Min and Max Key \(m_{i}\) and \(M_i\) through 
\begin{equation}
m_{i, d} = min(K_{i,d}), \ M_{i, d} = max(K_{i, d})
\end{equation}

where \(min(\cdot)\) and \(max(\cdot)\) are applied element-wise for each dimension \(d\). 

Given query \(q\), approximate attention score for block \(i\) is given by  
\begin{equation}
score_{i} = \sum_{j=1}^{d} max(q_j \times M_{i,j}, q_j \times m_{i,j})
\end{equation}

Then it selects Top-K blocks with highest \(score\) as sparse subset \(S\) for attention calculation.
\begin{equation}
S = \text{argtopk}(score, k)
\end{equation}

\textbf{DoubleSparsity} \cite{yang2024posttrainingsparseattentiondouble} approximates critical tokens efficiently through reducing the matrix multiplication dimension of calculating \(QK^T\) product. It first offline calculates outlier channels in \(QK^T\), denoted as \(C\). Then it selects Top-K tokens with highest approximate attention score \(\hat{s}\) as the sparse subset \(S\). 

\begin{equation}
Q_{label} = Q_{[C]}, \ \hat{s} = Q_{label}K^{T}_{label}, \ S = \text{argtopk}(\hat{s}, k)
\end{equation}

% Rectified Sparse Attention 
\textbf{ReSA} \cite{resa} combines training-free block-sparse estimation and GQA sharing, contributing to better efficiency. Besides, ReSA proposes a rectification stage to control the KV cache accumulation error. ReSA shows advantage on long-sequence generation tasks.

\subsubsection{Routing-based Block-Sparse Attention}

Routing-based Block-Sparse Attention learn the importance of each token block through trainable MLP layers, which act as gating network during inference to select critical blocks. 

\paragraph{Learnable Sparsity on Pretrained Models}

\textbf{SeerAttention} \cite{seerattention, gao2025seerattentionrsparseattentionadaptation} train the gating network on pretrained LLMs through a self-distillation manner. To obtain the importance score for each block, it first conduct pooling to \(Q\) and \(K\) along the sequence dimension, denoted as \(P_k\) and \(P_q\). The downsampled \(Q, K\) are then passed through a learnbale linear layer \(W_q\) and \(W_k\). Matrix multiplied results of projected \(W_q P_q(Q)\) and \(W_k P_k(K)\) go through the softmax operator as a gating process: 
\begin{equation}
score = \text{softmax}((W_q P_q(Q)) \cdot (W_k P_k(K)))
\end{equation}

The learnable linear layers are trained to align with the 2D maxpooled results of the original LLM through a self-distillation manner. The distillation loss is computed as:
\[gt = \text{MaxPool2D}(\text{softmax}(QK^T)),\ loss = D_{KL}(\,gt || score\,)\]

During inference, the gating score are used to predict block-level sparsity through Top-K or thresholding for sparse computation and efficiency.

\textbf{DSA}~\cite{deepseek3.2} inherits the self-distillation training pipeline from SeerAttention. However, to further alleviate the performance damage, DSA makes the sparsity computation on the token-level instead of block-level. Since the score computation of token-level sparsity is relatively high, DSA leverages smaller head dimension and lower precision to reduce the cost.  

\paragraph{Training-aware Sparse Attention}
\textbf{Landmark} \cite{mohtashami2023landmark} proposes using special landmark tokens to represent each block and trains the attention mechanism to directly retrieve Top-K blocks via these landmark tokens. However,  it did not experiment on large-scale pretrained models.

\textbf{MoBA} \cite{moba} integrate trainable sparse attention into the pretraining stage. It proposes Mixture of Block Attention, which applies the Top-K mechanism from MoE as gating mechanism to decide critical blocks for each query token. The importance score of each block is computed by the inner product between query token \(q\) and the mean pooling result of block \(K_i\) along the token dimension:
\begin{equation}
s_i = <q, P_{mean}(K_i)>
\end{equation}
Then the Top-K blocks with highest \(s\) score are selected for \(q\) in computing attention.

Notably, the Top-K block selection used by MoBA is not differentiable. Therefore, the sparsity pattern is still estimated in a training-free pattern in the pretraining stage, enabling both efficient inference and accelerated training. 

\textbf{NSA} \cite{nsa} introduces a training-aware mix-granularity sparse attention mechanism consisting of three branches \(C \in \{ \text{cmp}, \text{slc}, \text{win}  \}\),  which correspond to compression, selection, and sliding window strategies, respectively. NSA leverages a differentiable compression branch to learn the block selection score.

Combining the three branches, the attention output of NSA is given by

\begin{equation}
o = \sum_{c\in C} g^c \cdot \text{Attn}(q, K^c, V^c),\ g_c \in [0, 1]
\end{equation}

For the compression branch \(c = \text{cmp}\), block \(i\)'s key \(K_i \in \mathcal{R}^{d_k \times b}\) is compressed into a single key \(K_{i}^{\text{cmp}} \in \mathcal{R}^{d_k \times 1} \) through a learnable MLP layer \(\varphi\).  For the selection branch \(c = \text{slc}\), Top-K block are selected based on block importance score \(p\), which could be directly obtained from the compression branch. 
% For a simple case where the two branches share the same blocking scheme, block importance score \(p^{\text{slc}}\) is given by: 
% NSA sets selected block size to a multiply of the compression block size. The 
% block importance score for selected block \(i\) is the sum of compression branch's importance score \(p^{\text{cmp}}_j\) whose range is within this selected block: 
% \begin{equation}
% p^{\text{slc}}_i = \sum_{j \in i} p^{\text{cmp}}_j = \sum_{j \in i} \text{softmax}(qK^{\text{cmp}}_j)
% \end{equation}

% Select Top-k block with highest \(p^{\text{slc}}\). Selected block indices \(I\) is given by

% \[
% I = \{ i \mid \text{argtopk}(p^{\text{slc}}, k) \}
% \]

% Top-K block with highest \(p^{\text{slc}}\) are selected and used for computing attention output. 

\textbf{InfLLM-v2} \cite{minicpm4} adopts a training-aware Top-K block sparse attention mechanism similar to MoBA. To Top-K block selection accuracy, it divides blocks into small-granularity kernels with overlap and performs aggregation on kernel importance scores within each block. 
% \[
% K^{\text{slc}} = \text{Cat} [{K_i \mid i \in I}]
% \]

% \paragraph{Hardware-aligned algorithm Design}

% Blockwise selection follows the inherent distribution patterns of attention scores.

\subsubsection{System-level Design Choices}

Learning-aware Sparse Attention \cite{nsa, moba, minicpm4} begin to take kernel implementation and efficient execution into consideration. For efficient implementation of Block-Sparse Attention, FlashAttention \cite{flashattn2} is used for attention computation in an efficient tilling mechanism, introducing requirements and opportunies for better utilization of hardware resources, including: 
\begin{itemize}
    \item To avoid inconsistency in memory access, in SeerAttention \cite{seerattention} and MInference \cite{jiang2024minference}, block size \(b\) is typically set to a relatively large value of at least 64.
    \item To align with the minimal requirement of Grouped Matrix Multiplication instruction on GPU tensor cores, in NSA \cite{nsa} and InfLLM-v2 \cite{minicpm4}, the number of K, V heads within a query group are set to at least 16.
    \item To reduce memory access,  NSA \cite{nsa} and InfLLM-v2 \cite{minicpm4} forces sharing of selected blocks among query groups, which is done through conducting pooling on block-level importance score within query groups. 
\end{itemize}

% \paragraph{Kernel Design for Sparse Attention}

% \paragraph{Prefill/Decode }

% NSA: number of groups is 4, total of 64 attention heads 

% if token-level q, 16 凑 matmul 维度；else block-level q, 

\subsection{Clustering Attention}

Similar to Block-Sparse Attention, Clustering Attention aims to select the most critical tokens for decoding, but organizes tokens in data structures for better semantic property or sampling efficiency. 

% \textbf{Clustering Attention} groups tokens based on content or position similarity into  $K$ groups, $C_1, \dots, C_K$ and restricts attention within each cluster (i.e. $\mathcal{N}(i \in C_k) = C_k$). This approximates the dense attention matrix with a block-diagonal structure. Clusters can be predefined or learned during training.

% For example, Reformer\cite{reformer} uses Locality-Sensitive Hashing (LSH) to assign similar tokens to the same bucket. Routing Transformer\cite{routing} performs online $k$-means clustering per layer. ClusterFormer\cite{clusterformer} introduces a differentiable clustering module co-trained with downstream objectives. These methods reduce computation by grouping related tokens while maintaining performance through learned adaptability.

\textbf{RetrievalAttention} \cite{liu2024retrievalattentionacceleratinglongcontextllm} employs Approximate Nearest Neighbor Search (ANNS) for selecting critical K clusters. 
To address the challenge of the out-of-distribution nature between query and key vectors in the attention mechanism, it introduces an attention-aware vector search algorithm that adapts to the distribution of query vectors.

% \textbf{} also uses 

\textbf{ClusterKV}
\cite{liu2024clusterkvmanipulatingllmkv} select tokens at the granularity of semantic clusters, overcoming the issue of internal fragmentation of page-level retrieval methods such as Quest. After prefilling stage, 
tokens are clustered through K-means algorithm. The semantic similarity between token \(i\) and \(j\) are measured through the cosine similarity of key vectors \(\mathcal{D}(i,j) = 1 - \frac{<k_i, k_j>}{|k_i| \cdot |k_j} \).  Semantic clusters are represented by their centroids \( \mu_{1}, \mu_{2}, ..., \mu_{C} \in \mathcal{R}^{d}\). At each decoding step, clusters are selected based on the ranking of query token \(q\) and centroids \(\mu_i\)'s attention weights, i.e. \(q \mu_i^T\).

% 其实我不知道 RetrievalAttention 是怎么做的，如果他不是 cluster 可以改成引用 RetroInfer; 如果是 RetroInfer，

\textbf{MagicPIG} \cite{chen2024magicpiglshsamplingefficient} leverages Locality Sensitive Hashing (LSH) sampling to efficiently approximate attention computation. It employs LSH to map similar query and key vectors to the same hash buckets and offloads storage and partial computation to the CPU to address the KV cache bottleneck. It also introduces Oracle Top-K sampling as a better strategy than brute force Top-K.

\subsection{Bidirectional Sparse Attention}

Bidirectional Sparse Attention builds upon encoder-style architecture, using static pattern or block-level sparsity to accelerate attention computation.

Block sparsity is widely used in bidirectional sparse attention. 
BigBird \cite{bigbird} uses block-wise random attention, 
which acts as bridges to shorten indirect paths between tokens. 
Longformer \cite{longformer} uses static global-local hybrid attention. It also relies on block-level sparsity with additional global and random links, facilitating structured computation and memory-efficient parallelism.

Clustering-based methods are also used in bidirectional sparse attention. 
Reformer \cite{reformer} uses Locality-Sensitive Hashing (LSH) to assign similar tokens to the same bucket. Routing Transformer \cite{routing} performs online $k$-means clustering per layer. ClusterFormer \cite{clusterformer} introduces a differentiable clustering module co-trained with downstream objectives. These methods reduce computation by grouping related tokens while maintaining performance through learned adaptability.

%% file: sections/hybrid.tex
\section{Pretrained LLM with Efficient Attention}
\label{sec:plm}

\begin{figure}
    \centering
    \includegraphics[width=0.98\linewidth]{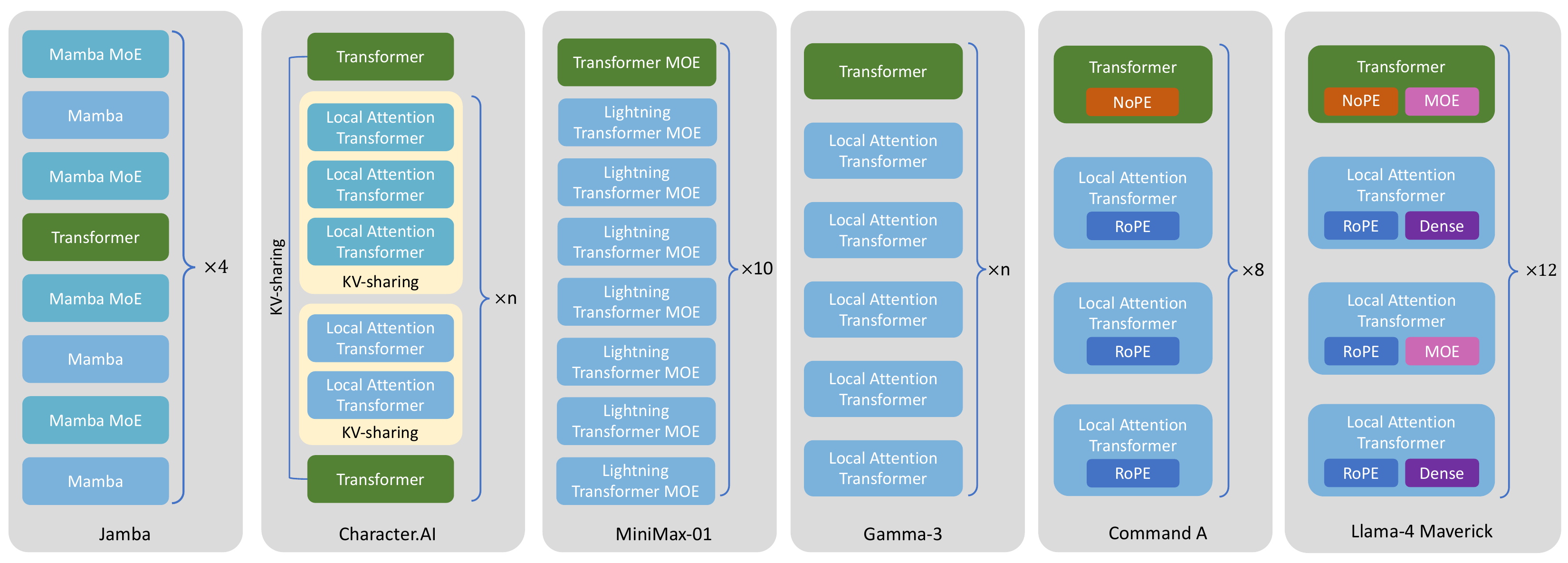}
    \caption{Architecture of different stacked models.}
    \label{fig:stacked_architecture}
\end{figure}

\subsection{Pretrained Models with Uniform Efficient Attention}
While early explorations of linear attention were often confined to smaller-scale models, recent advancements have demonstrated their successful scalability to the multi-billion parameter range, establishing them as viable and highly efficient alternatives to the standard Transformer. These models, built purely on linear attention or its architectural equivalents like State-Space Models (SSMs) and Recurrent Neural Networks (RNNs), retain their signature inference efficiency even at large scales.

\paragraph{RWKV-based model} The RWKV project represents a sustained and influential effort to create a scalable Recurrent Neural Network (RNN) architecture that combines the parallelizable training of Transformers with the efficient inference of traditional RNNs \cite{rwkv}.
% Its core design, equivalent to an element-wise linear attention with a time-decay mechanism, has proven effective at scale, culminating in models like the 7B-parameter EAGLE (RWKV-5)\cite{rwkv56}. The architecture's evolution has been marked by a continuous enhancement of model expressiveness while maintaining linear complexity. 
For instance, the EAGLE (RWKV-5) series introduced matrix-valued states to increase capacity, while subsequent iterations like Finch (RWKV-6) \cite{rwkv56} and Goose (RWKV-7) \cite{rwkv7} incorporated dynamic recurrence and expressive state evolution mechanisms (e.g., a delta-rule) to enable more complex, data-dependent state transitions.

\paragraph{Mamba-based model} The success of the Mamba architecture \cite{mamba}, with its data-dependent selection mechanism (Section \ref{sec:decay-data-dep}), has spurred a wave of adoption and scaling initiatives from major research labs.  \textbf{Falcon Mamba} \cite{falcon_mamba} is based on the pure Mamba-based architecture to demonstrate performance competitive with leading Transformer models on a wide range of general-purpose language benchmarks, validating the architecture's viability for such tasks while retaining its signature constant-time inference. Further evidence of this paradigm's potential is provided by \textbf{Codestral Mamba} \cite{codestral_mamba}, built on the Mamba-2 architecture. While specialized for code generation, it achieves state-of-the-art results on relevant benchmarks and supports a 256k token context, demonstrating the scalability and effectiveness of the SSM approach within a complex and structured domain.

\paragraph{Sparse-based model} MiniCPM-4~\cite{minicpm4} introduces a two-stage sparse attention mechanism that dynamically selects relevant key-value blocks for each query token based on semantic similarity. 
MiniCPM-4 leverages InfLLM-v2, a Block Sparse Attention variant to replace standard Attention mechanism.
% In the first stage, the model computes relevance scores between the query and a set of fine-grained semantic kernels using mean-pooled representations, then selects top-k blocks that intersect with high-scoring kernels.   Attention is subsequently restricted to all tokens within these selected blocks.   To reduce computation and memory cost, MiniCPM-4 adopts token-level granularity for queries and block-level granularity for key-value pairs, avoiding the inefficiencies of fully dense computation.   During inference, a block-sharing strategy is employed, allowing all heads within a group to attend to the same set of top-k blocks, further improving memory locality.
Moreover, a lightweight LogSumExp approximation enables efficient top-k selection, making the method scalable to extremely long sequences.   Together, these techniques allow MiniCPM-4 to balance fine-grained contextual awareness with tractable memory and compute requirements, making it a strong candidate for long-context modeling.

\subsection{Pretrained Models with Hybrid Efficient Attention}
With the increasing demand for efficient long-context modeling and diverse computational paradigms, recent research has extensively explored hybrid attention mechanisms. Such strategies combine global and local attention components, often interleaving specialized layers to balance computational cost and performance. The candidate model architecture illustrations are shown in Figure \ref{fig:stacked_architecture}.

\paragraph{Sparse Hybrid Model}
{GPT-3 \cite{gpt3} integrates a hybrid attention mechanism by interleaving dense and locally banded sparse attention layers, inspired by the Sparse Transformer \cite{sparsetransformer}. Dense attention provides full-context modeling, while sparse layers adopt fixed or strided patterns to reduce the number of attended tokens. This design enables GPT-3 to efficiently scale to large model sizes using a fixed context window of 2048 tokens, balancing modeling capacity and computational efficiency.

\paragraph{Linear-Full Hybrid Model}
Jamba \cite{jamba} and MiniMax-01} \cite{minimax01} combine linear and full attention layers to achieve an efficient trade-off between throughput and expressiveness. MiniMax-01 employs Lightning Attention across most layers, inserting Softmax-based full attention every eight layers. Jamba adopts a similar ratio, inserting one Transformer layer into every eight-layer Mamba block. Both achieve faster decoding and improved long-sequence performance by limiting the use of computationally intensive full attention.

\paragraph{Local-Full Hybrid Model}
Gemma 3 \cite{gemma3}, Command A \cite{rnope}, and LLaMA-4-Maverick~\cite{llama4} alternate between local and global attention layers, with a shared design philosophy of using global layers sparsely, e.g., every 4–6 layers, to enhance efficiency. While local layers adopt sliding-window patterns, the key difference lies in position encoding strategies. Gemma 3 modulates RoPE base frequencies—assigning 10K for local and 1M for global layers—to better capture long-range dependencies. Command A and LLaMA-4-Maverick mixes RoPE-based local layers with full attention layers that omit positional embeddings entirely, allowing stronger long sequence performance.

\paragraph{Advanced Hybrid Model}
Character.AI \cite{characterai} interleaves local attention with sliding windows and sparse global attention layers applied every six layers. Especially, they resue global attention layer's key-value representations across multiple non-adjacent layers. This KV sharing mechanism enables efficient long-context processing with reduced memory and latency overhead.

YOCO \cite{yoco} and Phi-4-mini-flash \cite{phi4miniflash} adopt a dual-decoder architecture that separates the prefill and generation phases. The Self-Decoder utilizes linear attention mechanisms such as RetNet and Sliding-Window Attention for both prefill and generation, while the Cross-Decoder is activated only during generation. A single-layer global KV cache is used throughout, allowing linear-time prefill and efficient decoding with minimal GPU memory consumption.

In summary, these recent advances underscore the trend toward hybridizing attention mechanisms to achieve balanced performance across varying computational constraints and sequence lengths. Each architecture uniquely contributes insights into effectively combining local detail management with global context integration, thereby providing valuable frameworks for future attention mechanism developments.

%% file: sections/conclusion.tex
\section{TAKEAWAYS}
\label{sec:takeaways}

Based on the current trajectory of both academic research and industrial deployment, several key consensus points emerge regarding the design of efficient attention mechanisms.

\paragraph{The Superiority of Hybridity} There is a growing consensus that while purely linear attention architectures offer impressive inference speeds, they frequently incur a non-trivial performance penalty in complex reasoning and ultra-long context retrieval. Consequently, hybrid models have emerged as the preferred architecture. These designs not only maintain the near-lossless performance of standard Transformers but demonstrate the potential to exceed them by balancing global associative memory with precise local retrieval.

\paragraph{The Gradient Bottleneck in Sparse Training} Training sparse attention models from scratch remains significantly more challenging than dense counterparts. This performance gap primarily stems from gradient sparsity, where restricted connectivity leads to suboptimal optimization dynamics and lower learning efficiency. While self-distillation has shown promise in bridging this gap, it remains an auxiliary solution, indicating that sparse-from-scratch strategies require further foundational innovation.

\paragraph{Complementary Strengths of Linear and Sparse Attention} From a unified functional perspective, linear and sparse mechanisms are increasingly viewed as complementary rather than competitive. In a modern efficient stack, linear attention serves as an effective state compressor for fast inference and KV cache reduction, while sparse attention acts as a high-fidelity module to replace full attention during decoding. A robust architecture leverages this synergy to navigate the Pareto frontier of computational throughput and predictive accuracy.

\paragraph{The Critical Role of Hardware-Algorithm Co-design} A recurring takeaway is that theoretical FLOPs reduction does not inherently translate to wall-clock speedups. Modern efficient attention is inextricably bound to hardware optimization. For instance, the utility of linear attention hinges on switching between recurrent and parallel forms based on the workload, while sparse attention's success depends on resolving the tension between theoretical sparsity and the massive parallelism required by modern GPUs.

\section{Outlook}
\label{sec:outlook}
This survey presents a comprehensive overview of efficient attention mechanisms, focusing on their algorithmic foundations, practical implementations, and integration into large-scale pre-trained language models. By categorizing linear and sparse attention into well-defined paradigms, we identify key design principles that enable scalability, computational efficiency, and long-context capability. We also analyze how these mechanisms are deployed in state-of-the-art models, either as standalone architectures or as part of hybrid designs that balance local and global computation.

Looking forward, we highlight several key directions that are expected to shape future research in this area:
\paragraph{Architectural Understanding of Hybrid Models} While prior work in linear attention has largely focused on standalone linear architectures, hybrid models are often constructed by combining off-the-shelf linear attention modules with dense or local components. However, it remains unclear whether a stronger linear backbone directly translates to improved hybrid performance. Future work should investigate hybrid models as a distinct architectural class, seeking to understand their composition, interaction effects, and optimization dynamics.
\paragraph{Lossless Sparse Attention and Extended Context} Sparse attention remains challenged by a trade-off between accuracy and computational gain. Fully trained sparse models often underperform dense ones, while post-training sparse approximations face limitations due to lack of end-to-end training. A major research frontier lies in developing sparse attention mechanisms that maintain the expressiveness and accuracy of dense attention, while scaling to much longer contexts. Additionally, the relationship between sparse budget and context length is poorly understood, where fixed top-k schemes may degrade with longer sequences, calling for more adaptive strategies.
\paragraph{Mechanistic Insights into Sparse and Hybrid Attention} While empirical studies have repeatedly demonstrated that hybrid attention models can match or exceed dense models using fewer attention computations, the underlying reasons for this effectiveness remain insufficiently explored. Besides, it is especially important to investigate whether the sparsity patterns that work well in synthetic benchmarks generalize to real-world tasks, and to characterize the limits of sparsity-based generalization.

As attention-based models continue to evolve, we expect further convergence between architectural innovation, theoretical insight, and hardware-aware design. We hope this survey provides a solid foundation for future research into efficient, high-performance language modeling systems.